\documentclass{cleiej}
\usepackage{graphicx}
\usepackage{hyperref}
\usepackage{doi}
\usepackage{lipsum}
\usepackage[utf8]{inputenc}

\usepackage{color}
\usepackage{graphicx}

\usepackage{amsmath}
\usepackage{amssymb}

\usepackage[linesnumbered,ruled]{algorithm2e}
\usepackage[noend]{algpseudocode}
\usepackage{booktabs}

\DeclareMathOperator*{\argmax}{arg\,max}
\DeclareMathOperator*{\argmin}{arg\,min}

\title{Towards Autonomous Reinforcement Learning: Automatic Setting of Hyper-parameters using Bayesian Optimization}

\author{
\bf Juan Cruz Barsce\\ 
Dpto. de Ing. en Sistemas de Información, \\ 
Fac. Reg. Villa María - UTN,\\
Villa María, Argentina\\
\it jbarsce@frvm.utn.edu.ar \\
\and
\bf Jorge A. Palombarini \\
Dpto. de Ing. en Sistemas de Información, \\
GISIQ, Fac. Reg. Villa María - UTN,\\
CIT Villa María - CONICET-UNVM, \\
Villa María, Argentina\\
\it jpalombarini@frvm.utn.edu.ar \\
\and
\bf Ernesto C. Martínez \\
Instituto de Desarrollo y Diseño \\
CONICET-UTN,\\
Santa Fe, Argentina\\
\it ecmarti@santafe-conicet.gob.ar
}

\begin{document}
\maketitle

\begin{abstract}

\noindent With the increase of machine learning usage by industries and scientific communities in a variety of tasks such as text mining, image recognition and self-driving cars, automatic setting of hyper-parameter in learning algorithms is a key factor for achieving satisfactory performance regardless of user expertise in the inner workings of the techniques and methodologies. In particular, for a reinforcement learning algorithm, the efficiency of an agent learning a control policy in an uncertain environment is heavily dependent on the hyper-parameters used to balance exploration with exploitation. In this work, an autonomous learning framework that integrates Bayesian optimization with Gaussian process regression to optimize the hyper-parameters of a reinforcement learning algorithm, is proposed. Also, a bandits-based approach to achieve a balance between computational costs and decreasing uncertainty about the \textit{Q}-values, is presented. A gridworld example is used to highlight how hyper-parameter configurations of a learning algorithm (SARSA) are iteratively improved based on two performance functions.  
\end{abstract}


\section{Introduction}

In recent years, with the notable increase of computational power in terms of floating point operations, a vast number of different applications of machine learning algorithms have been attempted, yet optimization of hyper-parameters is needed in order to obtain higher levels of performance. Such is the case of supervised learning algorithms like random forests, support vector machines and neural networks, where each one of them has its own set of hyper-parameters exerting influence on attributes such as model complexity or the learning rate of the algorithm, which are key issues in order to reap important benefits such as a better extrapolation to unseen situations, a reduction in computational times, a decrease in the model complexity, among others. Such hyper-parameters are often manually tuned and their correct setting makes the difference between mediocre and near-optimal performance of algorithms \cite{hutter_beyond_2015}. Several approaches have been proposed in order to optimize the hyper-parameters to reduce the error generated by a bad configuration of the model (e.g. \cite{bergstra_algorithms_2011, hutter_paramils:_2009, snoek_practical_2012, bengio_practical_2012, bergstra_random_2012}), with methods such as random search, gradient search, Bayesian optimization, among others.

In the particular case of reinforcement learning (RL) \cite{sutton_reinforcement_1998}, as opposed to supervised learning, there are not correct examples of desired behavior in the form of datasets containing a label for each vector of features, but instead a learning agent should seek for trade-off between exploitation with exploration so as to find a way of behaving using external hints. Given a state $s$ of the agent's environment, the RL problem is about obtaining the optimal policy (a function which, given sensed states, defines the action that should be chosen) in order to maximize the amount of numerical scores (rewards) provided by the environment. For the most commonly studied tasks that can be solved by using RL, where the agent may take a long sequence of actions receiving insignificant or no reward until finally arriving to a state for which a high reward is received (see \cite{mnih_human-level_2015}), the fact of having a delayed reward signal means larger execution times, so optimizing hyper-parameters using optimization strategies like Grid Search, Random Search, Monte Carlo or Gradient-based methods is not suitable for efficient autonomous learning. Alternatively, the Bayesian optimization strategy provides an approach designed to maximize the effectiveness of an expensive-to-evaluate function since it is derivative-free and less prone to be caught in a local minimum \cite{brochu_tutorial_2010}.

Among RL algorithms, such as Q-Learning \cite{watkins_q-learning_1992} or SARSA \cite{rummery_-line_1994}, there are several parameters to be configured prior to the execution of the algorithm, e.g. the learning rate $\alpha$ or the discount factor $\gamma$. In trivial environments such as a gridworld, the impact of a change in a parameter like the exploration rate $\epsilon$ of an $\epsilon$-greedy policy in an agent run is often easy to understand, but in the case of more complex environments and parameters such as the number of $N$ planning iterations in a Dyna algorithm \cite{sutton_integrated_1990,sutton_planning_1991}, the cut-off time of a given episode, or the computational temperature $\tau$ in a \textit{Softmax} policy, the impact that changes to these parameters may have in the agent's learning curve to solve a given task is normally unclear prior to its execution. Furthermore, when comparing the learning of different agents using different combinations of hyper-parameters, it is often unclear which one outperforms the other because it has superior learning rule, or simply because it has a better parameter setting. Moreover, if the environment is complex enough such that each learning episode is computationally expensive, having a bad parameter configuration could result in higher execution times that can eventually give rise to mediocre performance with a limited budget for experimentation\cite{hutter_paramils:_2009}.

In this work, a novel approach that employs Bayesian optimization to find a good set of hyper-parameters that improves the commonly used policies for a RL agent is proposed. In order to learn a near-optimal set of hyper-parameters using previous agent-environment simulated interactions, a Bayesian optimization framework is presented. In such an approach, a Gaussian process regression model is trained by using two functions that measure the performance of the agent in the simulated environment for unseen hyper-parameter settings. For meta-learning, a bandit-based algorithm is used to achieve a balance between the reduction of uncertainty about \textit{Q}-values and the computational costs of active exploration of untried action-state pairs. Finally, it is discussed an example that shows how the agent immersed in a gridworld iteratively converges to a near-optimal policy by querying the objective function for meta-learning of hyper-parameters which gives rise to learning episodes with increasingly improved sets of hyper-parameters. Results obtained demonstrate that autonomous reinforcement learning clearly outperforms a default configuration of the parameters in a RL algorithm. 

This work is an extended version of \cite{barsce_towards_2017} \footnote{Source code is available on-line at \url{https://github.com/jbarsce/rloptvb}.}, where a Bandits-based meta-learning layer has been added aiming to significantly reduce the computational expense involved in autonomously minimizing the variance of the different sequences of actions of a reinforcement learning agent - each with a different setting of hyper-parameters - to advantage. The proposed variance reduction is performed bearing in mind the fact that, in \cite{barsce_towards_2017}, most of resampling effort was invested in trying non-promising combinations of hyper-parameters (e.g. points whose metric is much lower than the average for alternative combinations), so performing additional simulations in those bad configurations is a waste of computational resources that can instead be directed at doing more simulations with promising combinations (e.g. at points where the acquisition function is maximized, as explained in section \ref{subsect:bayesian_optimization}). This is the main purpose that motivates the addition of another meta-learning component to decide, after evaluating the objective function at each point, if the outcome is good enough compared to others in order to resample the objective function. To this aim, new experiments have been performed, showing that a very significant reduction of computational times can be achieved with different meta-learning bandits-based approaches. However, for some bandit algorithms computational gains are not so important as one can expect, which is shown in Section \ref{subsect:optimizers_bandits}. Finally, in order to compare how the proposed approach performs against another common method for hyper-parameter optimization, a comparison between the convergence towards the optimum of the proposed algorithm against random search has been added, as it is shown in Section \ref{subsect:comparison_grid_random_search}.

The article is organized as follows. In Section \ref{sect:background} a revision of the different methods to be used later is presented, including reinforcement learning, Bayesian optimization, Gaussian processes, meta-learning algorithms and bandit-based meta-learning; such methods are discussed in the context of autonomous setting of a learning algorithm hyper-parameters. This is followed by a section devoted to the proposed framework of autonomous reinforcement learning, Section \ref{sect:rlopt_framework}, RLOpt, which aims to implement fast convergence to near-optimal agent policies while minimizing the number of queries in the objective function at the meta-learning level. This is followed by Section \ref{sect:case_study}, which presents the different computational experiments that were made in order to validate the proposed RLOpt approach. Finally, the work is summarized and the future research directions are presented in section \ref{sect:conclusion}.

\section{Background} \label{sect:background}

\subsection{Related Work} \label{sect:related_work}

The task of optimizing the hyper-parameters of machine learning algorithms is very challenging. A correct setting of hyper-parameters makes a significant impact on the effectiveness for the task being addressed, affecting the generalization errors in ways that are not easy to anticipate. Morever, only a small fraction of hyper-parameters are relevant to guarantee fast learning with few examples \cite{bergstra_algorithms_2011}. As the gradients are normally not available, most methods for setting hyper-parameters are based on optimizing them using pointwise function evaluations; this is often called black-box or derivative-free optimization. Alternatively, a common approach is manual tuning, which consists of setting the hyper-parameters arbitrarily and evaluating the generalization errors until the algorithm reaches some desirable performance for a given task. This is often used because it is fast (no algorithmic implementation is required) and sometimes the user can gain some empirical insights about the influence of hyper-parameters on learning method efficacy. However, hand-tuning of hyper-parameters is costly in terms of time and money, hinders the reproducibility of results obtained and limits machine learning to expert users, among other drawbacks \cite{hutter_paramils:_2009,bergstra_algorithms_2011}. As an alternative to hand-tuning, another common approach is grid search, which consists of training a model over every combination of a subset of the hyper-parameter space $\Theta$, in order to find a configuration $\theta \in \Theta$ that maximizes some metric. One of the problems of grid search is that it requires an amount of computational resources that increases exponentially with the number of hyper-parameters. As a result, a lot of poor hyper-parameter combinations are tried unecessarily \cite{bergstra_algorithms_2011}. In this regard, random search, an algorithm that consists on a random, independent sampling of the hyper-parameter space instead on a fixed sampling, is proved to outperform grid search \cite{bergstra_algorithms_2011}. As an alternative to gradient-free optimization in neural networks, a gradient-based hyper-parameter optimization method was proposed in \cite{maclaurin_gradient-based_2015}, that resorts to obtaining gradients by performing an exact reverse of stochastic gradient descent with momentum.

The main advantage of exhaustive search methods is that they are easy to run in parallel to easy the computation burden. However, they are based on samples that are independent from each other, and therefore they do not use the information of previous evaluations, so the convergence towards the optimum is slowed down. This is very significant when the budget of computational resources is scarce. As an alternative method aimed for expensive to evaluate functions, the sequential model-based optimization (SMBO) (a common algorithm is Bayesian optimization) consists of resorting to a probabilistic model in order to determine, according to the previous observations, what is the next point that yields, depending on the probabilistic model, the highest probability of optimizing an expensive performance function \cite{brochu_tutorial_2010, hutter_sequential_2011, shahriari_taking_2016}. Empirical evaluations highlight that SMBO performs better than random search on hyper-parameter tuning in some applications (see e.g. \cite{snoek_practical_2012}).

% comentado para reducir cantidad de citas y no hacer tan larga la sección Some recent hyper-parameter optimization frameworks are Optunity \cite{claesen_easy_2014}, Spearmint, which uses the algorithm proposed in \cite{snoek_practical_2012}, hyperopt \cite{bergstra_random_2012}, and AutoWEKA \cite{thornton_auto-weka:_2013}, among others. However, they are aimed at optimize the hyper-parameters on a supervised or unsupervised task.

Regarding hyper-parameter setting for reinforcement learning algorithms, common hyper-parameters are the step-size parameter $\alpha$, that regulates the speed of learning; the discount factor $\gamma$, that determines how future rewards are discounted (explained in section \ref{subsect:reinforcement_learning}), policy related parameters such as $\epsilon$ or $\tau$, that determines how the agent select its actions seeking to trade-off exploitation with exploration. In some algorithms, the trace parameter $\lambda$, that determines the decay rate of credit-assignment to previously visited states may be difficult to set. Each one of these hyper-parameters has been studied separately (for example, see \cite{sutton_step-size_1994}, \cite{dabney_adaptive_2012} and \cite{mahmood_tuning-free_2012} for the tuning of $\alpha$, \cite{yoshida_reinforcement_2013} for optimizing $\gamma$ and \cite{sutton_step-size_1994} and \cite{white_greedy_2016} for adapting $\lambda$). A meta-learning approach has been proposed in \cite{schweighofer_meta-learning_2003} to tune the hyper-parameters of a reinforcement learning for a problem related to animal behavior. The proposed approach is rather limited as the information to tune the hyper-parameters is obtained from dopamine neuron firing in animals. For high-level hyper-parameter setting in a reinforcement learning task, a relevant problem recently studied in this area is the problem of algorithm selection, where the hyper-parameter can be the learning algorithm itself (for example, \cite{laroche_reinforcement_2017}). Considering the scope and limitations of previously references in the field, our main motivation is to propose a general framework to autonomously tune all the hyper-parameters in a reinforcement learning algorithm so as to make hyper-paraneter setting part of the learning problem related to \textit{Q}-values.

% Comentado, partes descartadas por su posibilidad de extender demasiado esta sección
%Regarding the step-size parameter adaptation, as it is common method in machine learning algorithms. In particular, in reinforcement learning, \cite{dabney_adaptive_2012} \cite{mahmood_tuning-free_2012}

%Regarding the discount factor in reinforcement learning, \cite{yoshida_reinforcement_2013}

%Regarding some common policy dependent parameters such as $\epsilon$ and $\tau$, \textbf{TODO citar trabajos de RL que cambian epsilon y tau según avanza el algoritmo}

%Regarding the adaptation of the trace hyper-parameter, $\lambda$, it has been studied in \cite{white_greedy_2016}, where \textbf{TODO seguir desarrollando}.

\subsection{Reinforcement Learning} \label{subsect:reinforcement_learning}

As a method that can be combined with a large set of algorithms such as deep learning representations, as presented in \cite{mnih_human-level_2015}, and employed in applications such as self-driving cars \cite{shalev-shwartz_safe_2016} and in games such as Go \cite{silver_mastering_2016,silver_mastering_2017}, reinforcement learning (RL) is a form of computational learning where an agent is immersed in an environment $E$ and its goal is to converge to an optimal policy by performing sequences of actions in different environmental states and receiving reinforcement signals after each action is taken. At any discrete time $t$, the environment is in a state $s$ that the agent can sense. Each agent action $a$ is applied to the environment, making it to transition from state $s$ to a new state $s'$. A frequent setting is one that uses delayed rewards, i.e. by returning a non-zero reward whenever a goal state is reached and 0 otherwise. In such a setting, the agent could not normally determine whether an action is good or bad until a reinforcement is received, so the agent must try different actions in different states without a priori knowing the final outcome. For example, consider an agent learning to play Chess, where a check-mate situation would lead to a reinforcement of 1 or -1 depending on if it was a victory or a loss, whilst being in every other state would lead to a reinforcement of 0.

For the cases where the RL problem has a finite set of actions and states, and the sequences of $(state, action)$ satisfies the Markov property, then the problem can be formally defined as a Markov Decision Process (MDP) where $S = \{s_1, s_2, ..., s_n\}$ is the set of states, $A = \{a_1, a_2, ..., a_m\}$ is the set of actions and $ P_{a}(s,s')$ is the probability that the agent, being in state $s$, transitions to state $s'$ after taking an action $a$ and receive a reward $r$. An episode is defined by the finite sequence $s_0, a_0, r_1, s_1, a_1, \dots, a_{n-1}, r_n, s_n$. At each time step $t$, the agent senses its environmental state $s_t$ and selects the action $a_t \in A$ to take next using the (estimated) optimal policy. The latter summed up the experience gathered in previous interactions with the environment and tends to maximize future rewards.
To balance the search for short-term versus long-term rewards, future rewards are discounted by a factor $\gamma \in [0,1]$, such that the total reward $R_t$ at time $t$ is given by $R_t = r_t + \gamma r_{t+1} + \gamma^2 r_{t+2} + \gamma^3 r_{t+3} + ... + \gamma^n r_{t+n}$.

In order to solve a task using RL, the learning agent employs a \textit{policy} $\pi$ that defines, being in the environmental state $s$, what action must be taken in each given time $t$. With a given policy, the agent aims, for all states, to learn to maximize its \textit{action-value function} $Q_\pi(s,a)$, which represents the expected reward that would be obtained when starting from state $s$ by taking action $a$ and following policy $\pi$ afterwards. Given $s$, $a$ and $\pi$ the action-value function satisfies the \textit{Bellman equation}, expressed as

\begin{eqnarray}
Q_\pi(s,a) &=& \mathbb{E} (R_t \mid s_t = s, a_t=a) \label{bellman_equation}\\
		 &=& \sum_{s'} P(s' \mid s,a) (r(s,a,s') + \gamma V_\pi(s')) \nonumber
\end{eqnarray}
where $P(s' \mid s,a)$ is the probability that the agent transitions from state $s$ to state $s'$ after taking action $a$, $r(s,a,s')$ is the reward obtained by applying the action $a$ in state $s$ and transitioning to state $s'$ and $V_\pi(s')$ is the expected reward to obtain starting from state $s$ and by following policy $\pi$. Solving Eq. \eqref{bellman_equation} by applying iterations is what the different algorithm implementations of RL attempt to achieve. One of the most common algorithms is Q-Learning \cite{watkins_q-learning_1992}, which uses temporal difference at each time-step $t$ in order to update $Q(s,a)$. Q-Learning is an \textit{off-policy} algorithm, in the sense that each $Q(s,a)$ is updated considering the best value of $Q$ in the next time-step, regardless of the policy the agent actually uses. The update step is given by

\begin{eqnarray}
Q(s_{t},a_{t}) &\gets& Q(s_t,a_t) + \\
& & \alpha(r_{t+1} + \gamma \max_{a} Q(s_{t+1},a) - Q(s_{t},a_{t})) \nonumber
\end{eqnarray}

Other well-known reinforcement learning algorithms includes \textit{on-policy} algorithms such as SARSA \cite{rummery_-line_1994}, \textit{model-based} algorithms such as Dyna versions of \textit{Q}-learning and algorithms that incorporate \textit{eligibility traces} such as Q($\lambda$) \cite{peng_incremental_1996}, among many others. All of them have a fixed set of hyper-parameters such as the step-size $\alpha$, which impacts on the temporal difference update of $Q(s,a)$, or $n$, the number of times the model is used to simulate experience after each step as in Dyna-Q.

Regarding the agent's action-selection policy, a common one is $\epsilon$-\textit{greedy}, where the agent \textit{explores} actions that are a priori suboptimal, by taking a random action with probability $\epsilon$, and \textit{exploits} its current knowledge by taking the action that is believed to be the optimal one with probability $1-\epsilon$. Another common policy that the agent can follow in its learning policy is Softmax, where the agent takes an action with a probability that depends on its $Q(s,a)$ value. The most common implementation of this algorithm is the Boltzmann Softmax, which defines the probability of taking the action $a$ in state $s$ as 

\begin{eqnarray}
\label{eqn:softmax}
\pi(a \mid s) = \frac{\exp\{Q(s,a)/\tau\}}{\sum_{b} \exp\{Q(s,b)/\tau\}}
\end{eqnarray}
where $\tau$ is the computational temperature. By using this policy, the agent starts by employing an exploratory policy at the beginning with a"high temperature", switching over time to a more greedy policy as the value of the hyper-parameter $\tau$ is increasingly reduced, i.e. exploration "cools down".

\subsection{Bayesian Optimization} \label{subsect:bayesian_optimization}

As an approach typically used to maximize an expensive to evaluate function such as the one measuring the performance of a reinforcement learning agent in an uncertain environment, Bayesian optimization \cite{mockus_bayesian_1975} seeks to optimize an unknown objective function $f(X)$, where $X \in \mathbb{R}^n$, by treating it as a black-box function and defining a probabilistic model, a prior distribution, over its nature. Then, the Bayes' theorem is used to update the posterior belief about its distribution after observing the data $D_n = \{(X_1, f(X_1)), (X_2, f(X_2)), ..., (X_n, f(X_n))\}$ of new $f$ queries (the algorithm for BO is stated in Algorithm \ref{alg:bayesian_optimization} \cite{brochu_tutorial_2010}). In order to maximize the efficiency under a limited amount of possible queries i.e. a budget of queries for sampling $f$ , Bayesian optimization (BO) resorts to a cheap probabilistic surrogate model in order to determine the next point to query. This point is selected by maximizing a function called \textit{acquisition function}. To begin with, sampling methods such as Latin hypercube \cite{mckay_comparison_1979} are used. The acquisition function determines how much the agent exploits the current knowledge and selects points with the highest probability of being a new maximum of $f$ versus how much the agent explores a priori sub-optimal regions aiming to discover a reduced, yet feasible region of interest where the function maximum may be located with high probability. Emphasizing exploitation accelerates convergence, but the agent may often fall in a local optimum. On the other hand, emphasizing exploration ensures a more comprehensive sampling, but can significantly increase the time and cost needed to learn \textit{Q}-values. An acquisition function commonly chosen is the \textit{expected improvement function} \cite{mockus_application_1978,jones_efficient_1998}, because it provides a sensible trade off between exploration and exploitation by weighting the amount of improvement of a given point $X$ with regards to the current maximum $X^+$ by the probability that such point $X$ will improve over the current (known) maximum. The expected improvement acquisition function is given by

\begin{eqnarray}
\alpha_{EI} & = & \mathbb{E}(f(X)-f(X^+))  P(f(X) > f(X^+))  \\
& = & \Phi(Z) (\mu_n(X) - f(X^+)) + \phi (Z) \sigma_n(X) \nonumber
\end{eqnarray}
where $X^+$ is the point corresponding to the highest observed value of $f$, whereas $\Phi$ and $\phi$ are the cumulative distribution function and probabilistic density function of the standard Normal distribution, respectively, and $Z =\frac{\mu(X) - f(X^+)}{\sigma(X)}$.

\begin{algorithm}
    \SetKwInOut{Input}{Input}
    \SetKwInOut{Output}{Output}
    \Input{$\alpha_{model}$ - the acquisition function}
\For{t = 1, 2, ...}{
	$X_t = \argmax_X(\alpha_{model}(X \mid D))$\\
    Sample the objective function $y_t = f(X_t) + \epsilon$\\
    $D_{t+1} = add(D_n,(X_{t+1},y_{t+1}))$\\
    Update the statistical model\\
}
\Output{$\argmax_{X} f(X)$}
 \caption{Bayesian Optimization}
\label{alg:bayesian_optimization}
\end{algorithm}
In the Algorithm \ref{alg:bayesian_optimization}, it can be appreciated how, given an acquisition function, the objective function is queried in a controlled manner. The core sentence of the algorithm can be observed in line 2, where the acquisition function is optimized aiming to obtain the query point $X_t$ that, influenced by the chosen prior distribution, is the point whith higher probability of being the maximum of the objective function.

\subsection{Gaussian Processes} \label{subsect:gaussian_process}

In a supervised learning task, the agent is given examples in the form of correct feature-response pairs $D_n = \{(X_i, f(X_i))\}$, $i=1,\dots,n$ which are used to learn inductively how to make predictions for unseen $X$. Features or inputs are normally $X \in \mathbb{R}^D$, whereas responses can be either real valued, in such case the agent is learning an input-output \textit{regression} map, or categorically valued, i.e. $f(X) \in \{C_1, C_2, ..., C_n\}$. In the latter case the agent is learning a \textit{classification} task. Such feature/response pairs are fitted to training data, in such a way that the agent minimizes a loss function $L(f(X),\hat{f}(X))$ that measures the cost of predicting using the model $\hat{f}(X)$ instead of the actual (unknown) function, $f(X)$. A common loss function is the mean-squared error, given by

\begin{equation}
	MSE = n^{-1} \sum_i^n (f(X_i) - \hat{f}(X_i))^2
\end{equation}

Since the function $f(X)$ is unknown and expensive to query, a common approach is to assume that it follows a multivariate Gaussian distribution defined by a prior mean function $\mu_0(X) \to \mathbb{R}$ and a covariance function $k(X_i,X_j) \to \mathbb{R}$. Based on this assumption and by incorporating the pairs $D_n$ obtained from querying the objective function $n$ times, we are using a \textit{Gaussian process} \cite{rasmussen_gaussian_2008} as the surrogate model. A Gaussian process (GP) places a prior over the data set $D_n$, and, by applying the \textit{kernel trick} \cite{rasmussen_gaussian_2008}, it constructs a Bayesian regression model as follows: given $D_n$, the mean and variance of a new point $X$ is given by

\begin{eqnarray}
\mu_{n+1}(X) = \mu_0(X) + k(X)^T K^{-1}(Y - \mu_0)
\end{eqnarray}
\begin{eqnarray}
\sigma_{n+1}^2(X) = k(X,X) - k(X)^T K^{-1} k(X)
\end{eqnarray}
where $K$ is the covariance matrix, whose element $K_{ij} = k(X_i,X_j)$ and $k(X) = (k(X_1,X), ..., k(X_n,X))$ is a covariance vector composed of the covariance between $X$ and each of the points $X_i$ for $i=1,2,...,n$. On the other hand, the covariance function $k$ defines the smoothness properties of the samples drawn from the GP. The most common choice is the squared exponential function \cite{rasmussen_gaussian_2008}, given by

\begin{eqnarray}
k_{se}(X_i,X_j) &=& \sigma_f^2 \exp\{-\frac{1}{2} (X_i - X_j)^T \\ 
& &  diag(l)^{-2}(X_i - X_j)\} + \sigma_n^2 \delta_{ij}  \nonumber
\end{eqnarray}
where $\sigma_f^2$ is the variance of the function $f$, $\sigma_n^2$ is the noise signal, $l$ is a vector of positive values that defines the magnitude of the covariance and $\delta_{ij}$ is the Kronecker delta such that $\delta_{ij} = 1$ if $i=j$ and $0$ otherwise. This function is infinitely differentiable and it is often considered as unrealistically smooth for several optimization problems \cite{snoek_practical_2012}; an alternative is the Matérn covariance function. Assuming no prior knowledge about the hyper-parameters $\theta_{GP} = (\sigma_f^2, \sigma_n^2, l)$, a common approach is to maximize

\begin{eqnarray}
\label{eqn:log_likelihood}
\log  P (y \mid X,  \theta_{GP}) = &-& \frac{1}{2} (y-\mu_0)^T K^{-1} (y-\mu_0) \nonumber \\
&-& \frac{1}{2} \log | K | - \frac{n}{2} \log  2 \pi
\end{eqnarray}
where $y$ is the vector containing the observations $f(X_1), f(X_2), ..., f(X_n)$ and $\mu_0$ is the vector containing the prior means. Regarding $\theta_{GP}$, approaches such as Latin Hypercube Sampling or the Nelder-Mead optimization method can be used in order to obtain values that maximizes the likelihood of the dataset given the GP hyper-parameters.

\subsection{Meta-learning} \label{subsect:meta_learning}

The integration of algorithms for Bayesian optimization of parameters in a RL algorithm defines a more abstract layer of learning known as \textit{meta-learning} \cite{schmidhuber_simple_1996,thrun_learning_1998}. Meta-learning seeks to imitate the human capacity of fast generalizing concepts after seeing just a few examples, and is also referred to as \textit{learning to learn}. Meta-learning is closely related to transfer learning where an agent learns to perform better in a task by resorting to knowledge gain in successfully solving similar tasks \cite{wang_learning_2016}, even learning the learning algorithm itself \cite{lemke_metalearning:_2015}. Meta-learning is aimed at learning the specific bias of the task the agent is trying to do, i.e. the set of constraining assumptions that influence the choice of hypothesis for explaining the data. These biases allow the agent to gain prior knowledge that increase the efficiency in solving similar tasks \cite{lemke_metalearning:_2015}. Meta-learning involves at least two learning systems: a low level system that generally learns fast and an abstract learning layers that normally learns slower and whose purpose is to improve the learning efficiency at the low level system \cite{wang_learning_2016}. A well-known meta-learning example in machine learning is the boosting method \cite{freund_decision-theoretic_1997}, that consists on a set of "weak" learners that are trained sequentially and their predictions are weighted and combined to form a "strong" classifier that enjoys of a sensible higher prediction capability for class separation.

Regarding meta-learning approaches more specifically geared towards reinforcement learning, recent works include the incorporation of a general purpose RL agent as a meta-learning agent, used to train a recurrent neural network \cite{duan_rl$^2$:_2016, wang_learning_2016}. Also, a model-agnostic meta-learning layer that learns to generalize by performing a low number of gradient steps \cite{finn_model-agnostic_2017}, which can also be used in supervised and unsupervised learning has been proposed. Other meta-learning approaches have shown the effectivity of using Bayesian optimization at a more abstract learning layers, particularly in reinforcement learning for robotic applications (e.g. \cite{cully_robots_2015, pautrat_bayesian_2017}), where such layer is used to learn a policy for a damaged robot, updating the model after each iteration. This allows, for example, that the robot be able to walk by adapting its policy to a loss in motor functionality using knowledge acquired in previous faulty conditions.

\subsection{Bandit algorithms} \label{subsect:bandits}

The multi-armed bandit or \textit{ N}-Armed bandit problem is a decision problem in which a gambler must choose among \textit{N} arms or slot machines (also referred as "one-armed bandits"), where each arm has a latent probability distribution. The gambler seeks to maximize the value received from the slot machines, having a limited budget of pulls to be tried. The decision problem arises from the gambler having to decide between pulling those arms which rewarded a high payoff (exploitation) and pulling apparently less profitable arms where the value received was a lower value or unexplored arms that have not yet been pulled (exploration). This is a somewhat simplified reinforcement learning setting but without state related association of rewards received (it can be formalized as a one-state Markov decision process). This problem has several variants, but a common formulation consists of $N \in \mathbb{N^+}$ arms where $D = \{ \Phi_1(\mu_1,\sigma_1), ..., \Phi_N(\mu_N,\sigma_N) \}$ is the set of latent probabilistic payoff distributions corresponding to each $i=1, \dots, N$. At every discrete time-step $t=0, 1, \dots$, the gambler has to choose which arm $a_i(t) \in \{a_1(t), \dots, a_N(t)\} $ to pull, receiving in turn a reward $r \sim \Phi_i(\mu_i,\sigma_i)$, from which she may calculate the sample mean and the sample standard deviation, $\hat{\mu_i}$ and $\hat{\sigma}$, respectively. The problem of maximizing the sum of received rewards, $\sum r$, is what bandit algorithms attempt to solve. In this section there are listed some common ones (based on \cite{kuleshov_algorithms_2014}), which are also those used in this work.

Two common policies, as in the reinforcement learning problem, are $\epsilon$-greedy and Softmax. In the case of $\epsilon$-greedy, the policy for choosing an arm follows

\begin{equation} 
a_i(t) =	\begin{cases} 
                  \argmax_{a_{i=1,\dots,N}} \hat{\mu}_i(t) & \text{with probability } 1-\epsilon  \\
                  a_{i} & \text{with probability } \epsilon/N
   			\end{cases}
\end{equation}
This means that the best arm will be selected with a $1-\epsilon$ probability, and with $\epsilon$ probability a random arm is selected. A common variant of $\epsilon$-greedy is the greedy policy, which is the $\epsilon$-greedy policy with $\epsilon=0$ i.e. a policy which always selects the arm whose sampled mean is higher. On the other hand, the Softmax policy (as seen on Equation \ref{eqn:softmax}) states that each arm has a probability of being selected distributed in proportion to the sample mean, and given by

\begin{equation}
	P(a_{i} (t)) = \frac{\exp\{\tau \hat{\mu}_{i}(t)\}}{\sum_{b} \exp\{\tau \hat{\mu_{i}}(t)\}}
\end{equation}
Other common family of policies are the Upper Confidence Bound algorithms \cite{auer_finite-time_2002}, which consist in algorithms that maximizes the upper bounds of a confidence interval. Two algorithms used in this work are the UCB1 and the UCB1Tuned algorithms. In the UCB1 algorithm, the arm to be selected is determined in proportion of the sampled mean of that arm and a factor between an amount proportional to the amount of times each arm has been pulled and the current discrete time $t$. Analytically this is given by

\begin{equation}
	a_{i}(t) = \argmax_{a_{i=1, \dots, N}} \left[ \hat{\mu}_{i}(t) + \sqrt[]{\frac{2 \log{t}}{n_i}} \right]
\end{equation}
where $n_i$ is the amount of times the $i$-th has been pulled.
The UCB1Tuned algorithm is similar to UCB1, but it is extended by taking the variance into account by weighting the factor of UCB1 by an amount proportional to the sampled variance. Therefore, UCB1Tuned is given by:

\begin{equation}
	a_{i}(t) = \argmax_{a_{i=1, \dots, N}} \left[ \hat{\mu}_{i} + \sqrt[]{\frac{\log{t}}{n_i} \min \left[\frac{1}{4}, V_i(n_i) \right]} \right]
\end{equation}
where $V_i(n_i) = \hat{\sigma}_i^2 + \sqrt[]{\frac{2 \log{t}}{n_i}}$.

In this work, a meta-learning layer that combines a bandit learning algorithm and Bayesian optimization to control the performance, querying and selection of the hyper-parameters of a RL learning agent at a lower layer is proposed.

\section{RLOpt Framework} \label{sect:rlopt_framework}

The autonomous reinforcement learning framework proposed in this work, \textit{RLOpt}, consists of the integration of Bayesian optimization as a meta-learning layer in a reinforcement learning algorithm, in order to take full advantage of the agent's past experience with its environment given a different hyper-parameter configuration. The distinctive feature of RLOpt is the autonomous seeking for a good set of hyper-parameters that maximizes the efficiency of the agent learning an optimal policy without requiring any difficult-to-understand inputs from the user. In order to be able to optimize the outcome of each learning episode of an RL agent $A$, its interaction with a simulated environment given a set of hyper-parameters is treated as a supervised learning task at the meta-level of learning. In such a task, the set of hyper-parameters $\theta$ constitutes the input to a random function $f_A(\theta) \to \mathbb{R}$ which measures the influence of hyper-parameters on the learning curve. The output of such function is the performance metric in learning to solving the task. In other words, the hyper-parameters of the RL algorithm are taken as predictors of a real-valued performance metric for optimal policy learned so far, as in supervised learning; however, the distinctive aspect of the proposed framework is that it uses a supervised learning approach for hyper-parameters in a higher level of abstraction to improve to speed up the learning curve experienced by the agent while seeking to find in the best policy through interactions with the environment. Therefore, the objective of the RLOpt framework is resorting to data gathered $D_n = \{(\theta_i, f_A(\theta_i))\}$ as evaluative feedback to train a regression model, and then use Bayesian optimization to learn good algorithm configurations for $A$ with a minimum number of queries, i.e. learning episodes. On the other hand, RLOpt has its own set of hyper-parameters $\Theta$ at the highest level of abstraction which includes the number of queries that $A$ will perform and the parameters of the GP model such as the covariance function.

The framework consists of six components, as it is depicted in Fig. \ref{fig:bayesian_optimizer_framework}, separated by RL learning and a meta-learning layers. The components are described below:

\begin{figure}[h]
 \centering
 \includegraphics[width=0.60\textwidth]{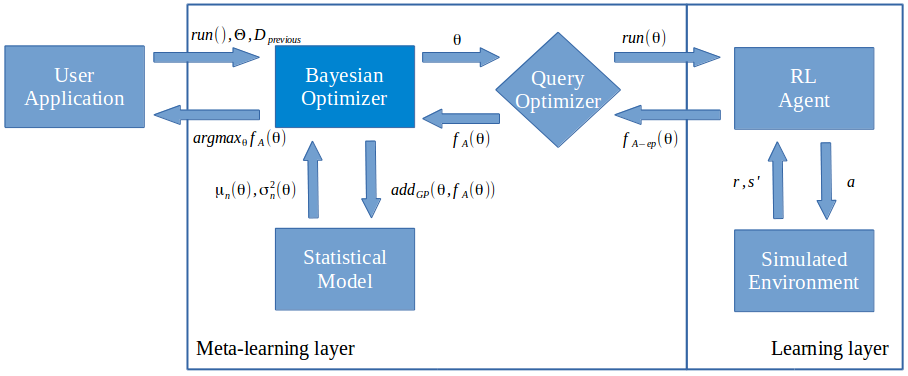}
 \caption{The Autonomous RLOpt Framework.}
  \label{fig:bayesian_optimizer_framework}
\end{figure}

\begin{enumerate}
\item The first component is a user application $U$, which sets the hyper-parameters $\Theta$ of the framework and adds previous agent-environment interaction pairs $(\theta, f_A(\theta))$, if they are available.

\item The second component is the Bayesian Optimizer module $BO$, which uses a statistical model and Bayesian optimization approach to decide what configuration is best to maximize the efficiency of the learning curve. The $BO$ module, according to $\Theta$, is set to run a certain amount of meta-episodes (i.e. queries to $f_A(\theta)$ which represents higher abstraction level episodes where several RL agent episodes are run under the same $\theta$ configuration).

\item The third component is the query optimization module $QO$, which stores all the pairs of the previous queries $(\theta, f_A(\theta))$ and, after receiving new $\theta$ vectors from $BO$, decides how many queries the agent will perform with the same $\theta$ configuration, in order to ensure that at least a small sample of queries in relevant regions are performed so as to reduce the impact of the stochasticity and to have a more accurate sample of the real value of $f$ for a given $\theta$. The purpose of the $QO$ component is to balance the trade-off between the precision of the $f(\theta)$ point and the computational cost of those queries.

\item The fourth component, is the reinforcement learning agent $A$, which receives configuration vectors from $QO$ and interacts with the environment, learning a policy by iteratively sensing the environment state and taking actions seeking a trade-off between exploitation and exploration. After receiving vectors of hyper-parameters, the agent starts each new simulation from having no prior knowledge about the policies found in previous instances with other vector of hyper-parameters. 

\item The fifth component is the environment $E$, which provides the simulation environment used to train the agent by receiving inputs from $A$ that can modify its state and returns rewards accordingly. 

\item Finally, the sixth component is the regression model (in this work a Gaussian process is employed) at the meta-level of learning which iteratively incorporates new data into its modeling dataset, whenever it is available, and use its current data in order to make a regression of the objective function at the level where hyper-parameters are learned. This regression model is employed to obtain a new configuration $\theta$ that maximizes the acquisition function so as to choose, hopefully, the next configuration of the agent learning $A$ where the maximum of $f_A$ has the greatest probability of being located, considering the previous meta-episodes. In the current implementation of the RLOpt, the statistical model employed is a Gaussian process; the main benefit of this model in the framework is that, by assuming a Gaussian prior about the data, it can learn after a few queries instead of needing a vast amount of training examples as methods such as deep learning models (see e.g. \cite{guo_deep_2016}). After receiving the signal from the $run()$ method initiated by the user $U$, the $BO$ proceeds to instantiate $A$, $E$ and the Gaussian Process as the statistical model.

\end{enumerate}

Based on prior experience, if it is previously provided by the user, the $BO$ module sets the configuration of $A$ and passes it to the $QO$ module, which, in turn, sets the configuration of $A$ and then sets an agent to run learning episodes for the first $\theta$ configuration. Then $A$ proceeds to successively apply actions which are defined by its current policy $\pi$ on $E$, sensing its next state and receiving a numeric reward. This process is repeated until a goal state or some stopping criterion is reached, and then the agent starts a new episode. After finishing a fixed amount of episodes, the results obtained by each episode of the agent are averaged in order to to calculate $f_{A}(\theta)$. Then, the average of the episode with the other outputs from different $\theta$ vectors is compared in the $QO$ module and, if the minimum number of queries for the same $\theta$ has been reached, a bandit algorithm is used to define if the current vector is good enough to query it again in order to see if whether it was a result of the stochastic effect. If that is assumed to be the case, then the same $\theta$ is used to query $f$ again. Otherwise, if the maximum amount of queries of the same vector has been reached, the current query is finished in the $QO$ and $f_A(\theta)$ is passed to the $BO$ so the configuration vector $\theta$, and the pair $(\theta, f_A(\theta))$ are added into the dataset used to fit the Gaussian process (statistical model). Once incorporated into the statistical model, the $BO$ starts using this new information to calculate $\mu_{n+1}$ and $\sigma^2_{n+1}$, in order to optimize the acquisition function so as to determine the next best configuration to perform in a new learning meta-episode. The optimizer then proceeds to initialize the agent with the new hyper-parameter configuration, repeating the cycle until a preset number of learning meta-episodes with their corresponding agent (lower-level) episodes have been made. The overall algorithm for the autonomous RLOpt Framework is depicted in Algorithm \ref{alg:bayesian_optimizer_framework}.

\makeatletter
\def\BState{\State\hskip-\ALG@thistlm}
\makeatother

\begin{algorithm}
\caption{RLOpt framework.} \label{alg:bayesian_optimizer_framework}
\SetKwInOut{Input}{Input}
\SetKwInOut{Output}{Output}
\Input{$\Theta$}

 \For{$n=1$ to $episodes_{BO}$}{
 	$\theta_{} \gets argmax_\theta(\alpha, \alpha_{opt})$\\
    $episodeQueries = 0$\\
    $nextQuery = True$\\
    \While{$nextQuery$}{
    	$episodeQueries = episodeQueries+1$\\
    	$init(A, episodes_A)$\\
        $f_{A-avg}(\theta_{}) \gets 0 $\\
        \For{$ep=1$ to $episodes_A$}{ 
        	$restart(A)$\\
            $run(A \mid \theta)$\\
            $saveExecution(A)$\\
            $f_{A-avg}(\theta_{}) \gets f_{A-avg}(\theta_{}) + f_{A-ep}(\theta_{})$\\
        }
    	$nextQuery = decideIfNextQuery(A)$\\
    }
    $f_{A-avg}(\theta_{}) = f_{A-avg}(\theta_{})/(episodeQueries * episodes_A)$\\
	$add_{GP}(\theta_{}, f_{A-avg}(\theta_{}))$\\
 }
 
\Output{$\argmax_\theta f_{A-avg}(\theta)$}
\end{algorithm}

In Algorithm \ref{alg:bayesian_optimizer_framework}, $\Theta$ is the framework configuration vector. This vector is composed by $A$, which is the RL algorithm; $\mu_0$ which is the prior mean vector; $\sigma_f^2$, the noiseless variance of $f$; $\sigma_n^2$, the noise level of the GP; $l$ the vector that defines the magnitude of the GP covariance function; $\alpha$, the acquisition function; $\alpha_{opt}$ which is the function that optimizes $\alpha$; the cut-off time $k$; $minRuns_\theta$ and $maxRuns_\theta$ which are the minimum and maximum amount of learning episodes to be made under the same configuration $\theta$, respectively; $init_{LH}$, the amount of optional training meta-episodes used to add information sampled by a Latin hypercube to the covariance matrix of the GP, and finally $episodes_{BO}$, which is the number of the $BO$ meta-episodes to run.
On the other hand, the random function $f_A(\theta)$ is sampled after executing a fixed amount of agent episodes given by $episodes_A$ and under a given configuration vector $\theta$. After initializing the optimizer, the RL algorithm of $A$ that runs the simulation and the acquisition and covariance functions are instantiated. If the $BO$ has not prior query points, before the beginning of the first episode the covariance matrix is initialized by sampling a random number of $init_{LH}$ points in order to start the first set of GP model predictions with non-trivial values of mean and variance (this step is skipped if $init_{LH}=0$). Whenever the $BO$ changes the hyper-parameters of $A$ and resets its knowledge, it also resets $E$ by returning it to its initial state, so as to each agent can interact with the environment in similar conditions.

Because $A$ involves a stochastic nature in its decision making policy to make room for exploration, the result of the performance metric $f_A(\theta)$ may vary from one simulation run to another even with the same configuration $\theta$. In order to minimize its variance and obtain a more significant value of $f(\theta)$ for a given $\theta$, $f(\theta)$ can be queried several times with the same $\theta$ and the result of the queries of the objective function is averaged over all queries made with the same vector. The core method where the $QO$ module decision is performed is the $decideIfNextQuery$ function, that returns $True$ if it has been decided that the best choice is to query the objective function with the same configuration, and $False$ otherwise. In order to make this decision, a N-armed bandit algorithm is employed such that, at any time, there are two arms that can be chosen: the first is an arm that represents the decision to not keep querying the objective function with the same configuration, because the sampled mean and the variance of the $f(\theta)$ samples with the current $\theta$ vector are not good enough, in comparison to results obtained using other configuration vectors. Likewise, the other arm represents the decision of performing another query to the objective function with the same configuration, because the mean and variance of $f(\theta)$ under the same $\theta$ are sufficiently good compared to the other values of the performance metric.

On the other hand, the method $init(A, episodes_A)$ sets the agent to its initial environmental state, thus erasing its previous knowledge. This is done for two reasons: firstly, because previous knowledge relied on hyper-parameters which are different to the current set, and therefore the new knowledge is biased in unpredictable ways; secondly, in order to give the agent a new unbiased configuration to interact with the environment so as to evaluate how the learning curve converges when starting with no prior knowledge. On the other hand, the method $restart(A)$, used to start a new agent episode, restores the agent to the initial state of the environment, keeping its acquired knowledge. The procedure $run(A \mid \theta)$ runs the agent episodes under $\theta$ in its environment from its initial state and until $A$ reaches a final state or the amount of steps corresponding to the cut-off time. The method $saveExecution(A)$ stores the results obtained in the episode, $f_{A-ep}(\theta)$, for further use, e.g. by saving the amount of steps required for the agent to reach the final state. Finally, when the last episode of $A$ under the same configuration $\theta$ finishes, the $BO$ saves and averages the results obtained, attaching them to the dataset $D$, incorporating this new observation in the statistical model. The cycle repeats itself until a certain amount of $BO$ meta-episodes are completed, and then the configuration $\theta$ that maximizes $f_A$ is returned.

In order to compare the efficiency between an agent $A$ against another agent $A^{\prime}$ in how both performs when taking actions to solve a task, two performance metrics were integrated into the framework. The first is the average amount of time steps per episode, 

\begin{equation}
f_A(\theta) = \frac{\sum_i^{n_{ep}} t_{i}}{n_{ep}}
\end{equation}
Where $t_i$ is the amount of time steps employed in the episode $i$, and $n_{ep}$ is the total number of episodes experienced by $A$ under configuration $\theta$. This metric aims to reward the agent that solves the task in the minimum possible time, i.e. taking the minimum possible amount of time steps to reach the goal state, where $n_{ep} \leqslant f (\theta) \leqslant kn_{ep} $. That is, if $A$ cannot reach its objective before the cut-off time, the amount of steps employed in that episode will be added. When using this measure, the framework will treat the optimization as a minimization problem by searching for $\argmin_\theta f_A(\theta)$.

On the other hand, the second metric used is the amount of \textit{successful runs} per episode. The following metric adds a notion of success to the agent episodes by making a distinction between successful and unsuccessful episodes of the agent, instead of measuring the performance by a numerical score. For an episode to be considered as a success, it considers an objective that must be accomplished regardless of any other considerations such as the amount of steps the agent required to reach the goal state. The metric is defined by

\begin{equation}
f_A(\theta) = \frac{\sum_i^{n_{ep}} s_{i}}{n_{ep}}
\label{eq:episode_per_step_optimizer}
\end{equation}
Where, for each episode $i$, $s_i = 1$ if that episode was a success; $s_i = 0$ otherwise. The idea behind this performance metric is to adapt to problems where, being the unknown objective function is expensive to query, such as the function $f_A(\theta)$ defined above. Thus, it is more important for the learning agent to succeed while experiencing the episode rather than maximizing some other performance metric. For example, consider an agent which uses reinforcement learning in order to learn how to better perform rescheduling tasks. For such an agent, it is mandatory to find a repaired schedule which is feasible in minimum time rather than finding the best schedule that is both feasible and optimum in some sense, e.g. having a minimum total tardiness\cite{palombarini_generating_2014}.

\section{Case Study} \label{sect:case_study}

\subsection{Experimental Setup} \label{subsect:experimental_setup}

\begin{figure}
 \centering
 \includegraphics[width=0.60\textwidth]{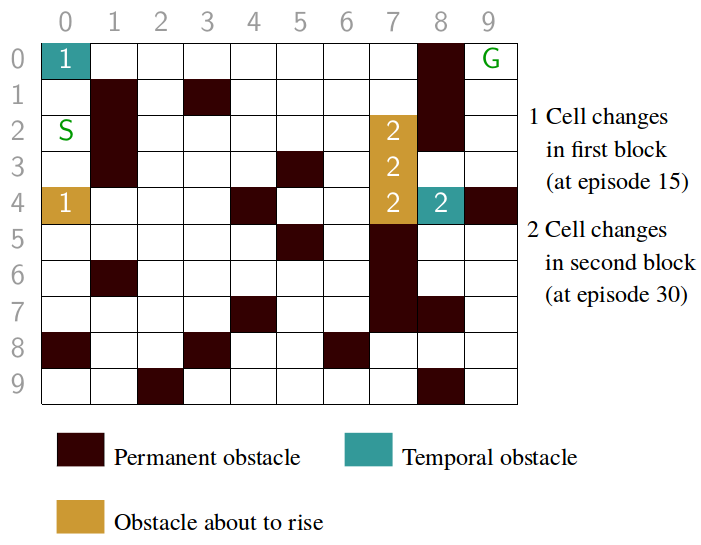}
 \caption{Double blocking gridworld environment.}
  \label{fig:double_blocking_gridworld}
\end{figure}

In order to illustrate the proposed approach, results obtained for a gridworld problem are presented. Based on the Blocking Maze example in \cite{sutton_reinforcement_1998}, the proposed example consists of a grid where an agent starts from an initial state $S$ aiming to reach a final state $G$, in which the environment returns a reward of 1 (as opposed to transitions to any other state where the reward obtained by the learning agent is always 0). There is a set of obstacles between them, as depicted in Figure \ref{fig:double_blocking_gridworld}. Such obstacles are either permanent obstacles where the agent cannot pass, obstacles that can be temporarily overcome for the agent, but at a certain episode they become permanent obstacles, and temporal obstacles where the agent will be able to pass after some amount of episodes have elapsed. There are two instances where the environment may change in each agent simulation: the first happens whenever an agent instance reaches episode 15 and the other when the same agent reaches the episode 30. The idea behind those environmental changes is to test how the agent adapts to a somehow different environment with a given configuration of parameters.

In this work, two variants of the framework were employed: the first one which uses as performance metric the average amount of time steps per episode, and the second one which uses as metric the amount of successful steps per episode, which, in this example, is the amount of episodes where the agent reached the final state (i.e. in less time-steps than the cutoff time). The algorithm selected for the agents to perform the different experiments was SARSA($\lambda$), which is a variant of the on-policy SARSA algorithm that uses eligibility traces, a mechanism to propagate the $Q(s,a)$ values to the state-action pairs visited in the past, proportionally to how much time steps have elapsed since the last time the agent visited each pair. The algorithm, in turn, is run under an $\epsilon$-greedy policy and with a cut-off time of 400 time-steps. In this particular setting, each agent instance $A_i$ has as the configuration vector $\theta_i = (\alpha_i, \epsilon_i, \gamma_i, \lambda_i)$, where the parameters represents the learning rate, the exploration rate, the discount factor and the eligibility traces decay rate, respectively. Regarding the statistical model configuration, the GP hyper-parameters were not optimized after each step in order to use the same configuration between all the simulations. Instead, the following hyper-parameters, which were determined empirically, were used: $\sigma_f^2 = 0.8$, $\sigma_n^2 = 0.17$, $\mu_0 = (0, 0, 0, 0)$, $l = (-0.12, -0.12, -0.12, -0.12)$, expected improvement as the acquisition function and squared exponential as the covariance function. The hardware configuration used to run the different experiments consisted in a computer with 8 GB of RAM and 4 x 3.20 GHz processors.

Three different set of experiments were run in order to validate RLOpt. In the first experiment set, the focus is to validate how the framework iteratively seeks for a $\theta$ vector that optimizes a performance metric $f$ in 30 meta-episodes considering two alternative ways for measuring it. On the other hand, the second experiment set uses the best two vectors of $\theta$ for each metric of $f$ found by the optimizers performing a simulation, showing how their corresponding learning curves converge in comparison with a default $\theta$ configuration. Finally, in the third experiment set, an extension of the first experiment is presented, that illustrates how adding a N-armed bandit algorithm in the meta-learning layer makes room for a considerable reduction in the amount of queries with the same theta and, at the same time, maintains a similar or better convergence for both performance metrics. The experiments and their corresponding results and analysis are discussed in the subsections below.

\subsection{Convergence of the optimizers towards the optimum of the objective function} \label{subsect:optimizers_convergence}

\begin{figure}
 \centering
 \includegraphics[width=0.60\textwidth]{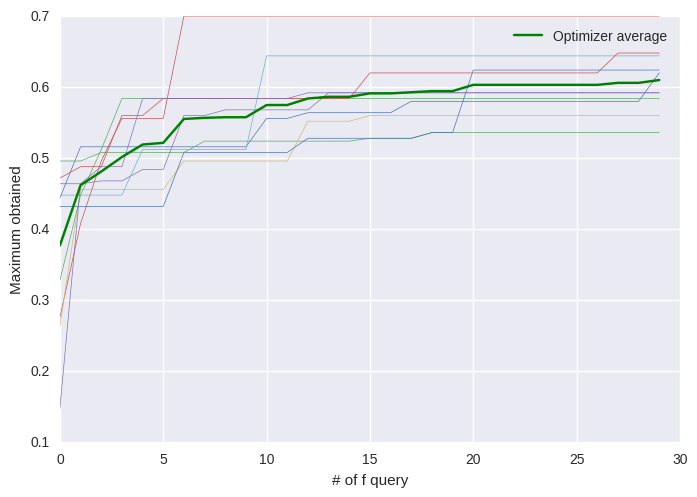}
  \caption{Convergence towards the maximum number of successes per episode.}
  \label{fig:convergence_towards_the_maximum_of_the_success_measure_optimizer.png}
\end{figure}

\begin{figure}
 \centering
 \includegraphics[width=0.60\textwidth]{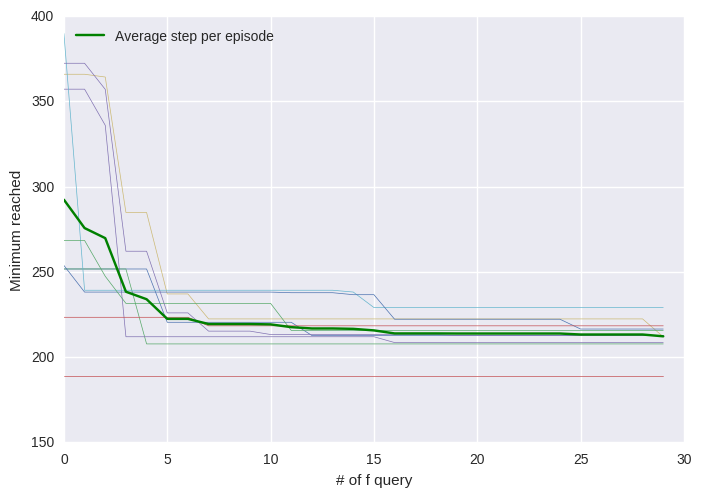}
  \caption{Convergence towards the minimum number of steps per episode.}
  \label{fig:convergence_towards_the_minimum_episode_per_step_optimizer.png}
\end{figure}

In this set of experiments, the two variants of RLOpt were executed 10 times in order to compare how fast they converge towards the maximum value of $f$, where each execution consisted of 30 different queries to the performance function. For each query, a RL agent with a given configuration $\theta$ is engaged in running 5 times 50 episodes, in order to obtain the average of the query. Each execution of the success metric variant took an average of 2:12 hours, whereas the execution of the variant with the step-per-episode metric took an average execution time of 2:39 hours. The two variants of the framework were run in parallel, where the ten executions were run in batch, having finished in about 26:30 hours. The results of the optimization for the number of successes and the number of steps per episode are depicted in Fig. \ref{fig:convergence_towards_the_maximum_of_the_success_measure_optimizer.png} and Fig. \ref{fig:convergence_towards_the_minimum_episode_per_step_optimizer.png}, respectively, where each curve represents how, for the success metric and steps per episode metric, the maximum and minimum that were found for each execution changes for a given meta-episode, and the thicker curve represents the average optimum value found for all runs made using the same framework configuration alternative.

Regarding the variant with the success metric, as can be seen in Fig. \ref{fig:convergence_towards_the_maximum_of_the_success_measure_optimizer.png}, the autonomous learning agent is able to found iteratively the average maximum percentage of success per episode starting from an average of near 0.3772 and converging to a maximum of 0.61 after 30 queries, which represents a 61.7\% increase in the success rate from the initial to the final query, with about a 2.12\% average increase of the maximum for each query. The learning curves reached a maximum of 0.7 successful episodes per episode and the worst case was 0.536 successful runs per episode. On the other hand, regarding the convergence, it can be appreciated in various learning curves and in the average curve that this variant is prone to be caught in a local maximum: there were a total of 12 meta-episodes where not a single of the 10 runs increased its maximum value at all. Another noticeable aspect is that the major increase of the maximum happened between the 2th and the 7th query on all the runs made, where the average increase of the maximum of all the queries made in that interval is 6.88\%; and after that interval the average increase is reduced to 0.41\%.

On the other hand, regarding the variant using the number-of-steps-per-episode metric displayed in Fig \ref{fig:convergence_towards_the_minimum_episode_per_step_optimizer.png}, the  minimum of the average amount of steps per episode starts in 276.4 in the first query, converging after 30 queries to a minimum of 211,7, which is a decrease of about 23\% of the initial average steps per episode, with an average of 0.80\% decrease of the minimum of each run; the best minimum value for a single run reached 188.95 average-steps-per-episode while the worst minimum value is 229.06 average-steps-per-episode. Regarding the convergence of each execution, it can also be observed that in the different curves the optimizer is prone to stuck in a local minimum. The number of episodes where in no one of the 10 executions is able to change its minimum value using this metric which is 14. As in the other variant, a remarkable aspect is that, between the 2th and the 6th query of all the runs, a worth noting decrease is produced in the minimum of the majority of the curves, where various of the minima can be seen. The average decrease of the minimum found in this period is 3.6\%. After that gap, the convergence slows down to an average decrease of 0.19\% per episode. In both performance metrics used this fact can be interpreted as that the first few queries are the most important to determine the optimum of each execution, thus suggesting that a minor amount of meta-episodes could have been used instead of 30 to save computational time while obtaining similar results.

\begin{figure}
 \centering
 \includegraphics[width=0.60\textwidth]{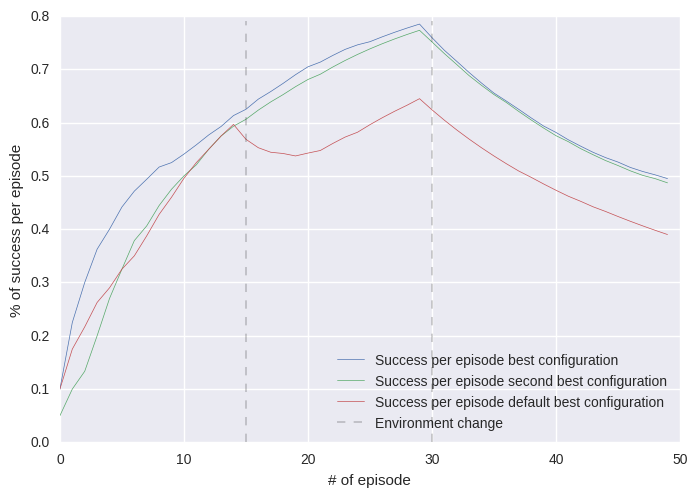}
  \caption{Learning curve of the variant with the success metric.}
  \label{fig:Learning_curve_success_Optimizer.png}
\end{figure}

\begin{figure}
 \centering
 \includegraphics[width=0.60\textwidth]{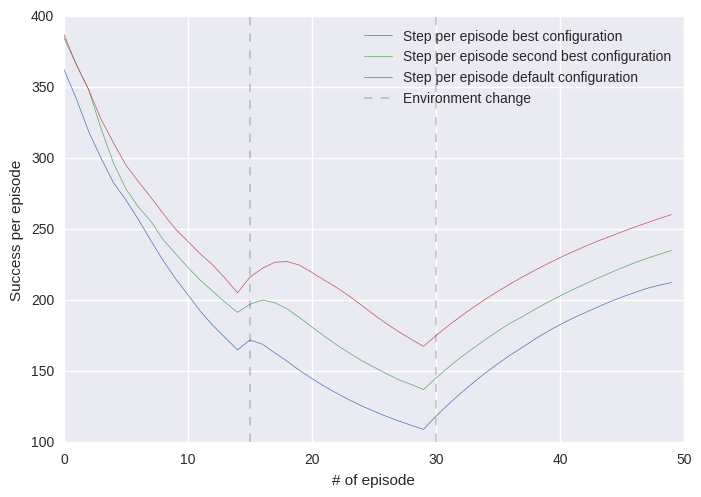}
  \caption{Learning curve of the variant with the step-per-episode metric.}
  \label{fig:Learning_curve_episode_step_Optimizer.png}
\end{figure}

\subsection{Comparison of Learned Policies with a Default Configuration} \label{subsect:comparison_learning_with_default_config}

In order to visualize how is the learning of the different agents that use the global optimum obtained from each variant, and how they compare to a default configuration used in RL, the best two configurations for each alternative performance metric had been run again 20 times in a RL agent and averaged in order to compare them with a default configuration for RL agents. For such default configuration, the parameters selected as default were the default parameters employed for a SARSA($\lambda$) agent in the Soar Cognitive Architecture \cite{laird_soar_2012} (specified in the Soar manual), that consisted on $(\alpha = 0.3, \epsilon = 0.1, \gamma = 0.9, \lambda = 0.001)$. The results can be seen in Table \ref{table:results_success_optimizer} and Table \ref{table:results_episode_step_optimizer}, whereas the convergence, represented by a learning curve that averages all the previous episodes, can be appreciated in Fig. \ref{fig:Learning_curve_success_Optimizer.png} and Fig. \ref{fig:Learning_curve_episode_step_Optimizer.png} for the variant with the success metric and for the variant with the step-per-episode metric, respectively.

\begin{table}
\centering
\caption{Hyper-parameters and results for the success metric variant}
\label{table:results_success_optimizer}
\begin{tabular}{@{}lcc@{}}
\toprule
                           & Hyper-parameters 					   & Metric \\ \midrule
Best Configuration         & $(0.538, 0.49, 0.69, 0.686)$          & 0.495                        \\ \midrule
Second Best Configuration  & $(0.582, 0.553, 0.653, 0.321)$        & 0.487                        \\ \midrule
Soar Default Configuration & $(0.3, 0.1, 0.9, 0.001)$              & 0.39                         \\ \bottomrule
\end{tabular}
\end{table}

\begin{table}
\centering
\caption{Hyper-parameters and result for the steps-per-episode variant}
\label{table:results_episode_step_optimizer}
\begin{tabular}{@{}lcc@{}}
\toprule
                           & Hyper-parameters               & Metric \\ \midrule
Best Configuration         & $(0.607, 0.191, 0.667, 0.707)$ & 212.49  \\ \midrule
Second Best Configuration  & $(0.291, 0.38, 0.5, 0.784)$    & 235.07  \\ \midrule
Soar Default Configuration & $(0.3, 0.1, 0.9, 0.001)$       & 260.29  \\ \bottomrule
\end{tabular}
\end{table}

When the performance metric is the success, it can be appreciated that both agents with its configurations obtained as a result of the optimization perform similarly. Moreover, both behave better than the default configuration, which in turn start converging similarly to the second best configuration but their adaptation is significantly worse compared to the first environmental change. The best configuration starts consistently better than the other two, whereas their average is significantly decreased after the second environmental change, in a similar profile in comparison to the other two runs. On the other hand, regarding the steps-per-episode variant, the agent with the default configuration again performed worse compared to the agent with the second best configuration, despite it starts with a poor performance in comparison to the agent with the default configuration. Compared with the other variants, there was a clearer distinction in this metric between the performance of the different agents, where at least 20 steps per episode separated all three runs and this separation increases after the first environmental change.

\subsection{Comparison with random search} \label{subsect:comparison_grid_random_search}

In order to assess how the results obtained in the precedent section compares with other hyper-parameter optimization methods, convergence curves obtained used the random search (RS) method help providing a representative example (as described in Section \ref{sect:related_work}). An implementation of the RS algorithm was run with the same budget and configuration used in RLOpt: 10 executions of the algorithm, where each execution consists of 30 meta-episodes. The ten executions were set to run twice, one for each performance metric.

\begin{figure}
 \centering
 \includegraphics[width=0.60\textwidth]{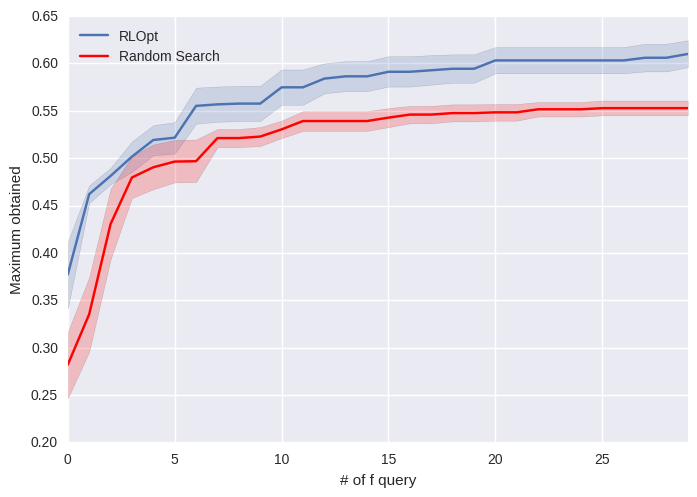}
  \caption{Comparison of RLOpt convergence to the maximum against the convergence of random search (success metric).}
  \label{fig:Comparison_grid_random_search_success.png}
\end{figure}

\begin{figure}
 \centering
 \includegraphics[width=0.60\textwidth]{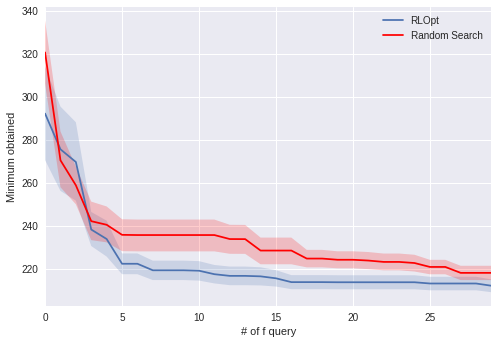}
  \caption{Comparison of RLOpt convergence to the minimum against the convergence of random search (steps-per-episode metric).}
  \label{fig:Comparison_grid_random_search_stpep.png}
\end{figure}

The results of the convergence can be seen in Fig. \ref{fig:Comparison_grid_random_search_success.png} and Fig. \ref{fig:Comparison_grid_random_search_stpep.png} for the success and the steps- per-episode metric, respectively. The blue and red curves represent the observed behavior for RLOpt and Random Search, respectively. In each learning curve, the solid line in the middle corresponds to the mean of ten independent executions, whereas the lighter area around that line has drawn based on the standard deviation and confidence intervals. As can be seen in both curves, RLOpt consistently tends to find better optima conbinations of hyper-parameters, on average, than RS. More specifically, for the success metric, RLOpt consistently increases the distance with respect to RS. on the other hand, for the step-per-episode metric, RLOpt does a better start than RS in the first 5 meta-episodes. Later on, RS starts to reduce its distance to RLOpt from the 11th meta-episode and onwards, reaching the minimum distance at the 28th meta-episode.

%\textbf{TODO poner porcentajes?}

\subsection{Variants with \textit{N}-Armed Bandits Decision Algorithms} \label{subsect:optimizers_bandits}

In order to reduce the total amount of times $f(\theta)$ is queried, and therefore execution times, the experiments presented in this section are based on the experiments made previously in Section \ref{subsect:optimizers_convergence}, extending them by adding executions with \textit{N}-armed bandits for each variant of the optimizer. Five variants of the bandit-based algorithms presented in Section \ref{subsect:bandits} were used: $\epsilon$-greedy with $\epsilon=0.2$, greedy (i.e. $\epsilon$-greedy with $\epsilon=0$), Softmax with $\tau=1$, UCB1 and UCB1Tuned. On each execution of the optimizers, a setting of $minRuns_\theta = 2$ and $maxRuns_\theta = 5$ were used, so each optimizer run would at least query $f$ twice for each $\theta$, and at most 5 times in order to avoid impasses whenever a local optimum is found and thus saving computational resources.

\begin{figure}
 \centering
 \includegraphics[width=0.60\textwidth]{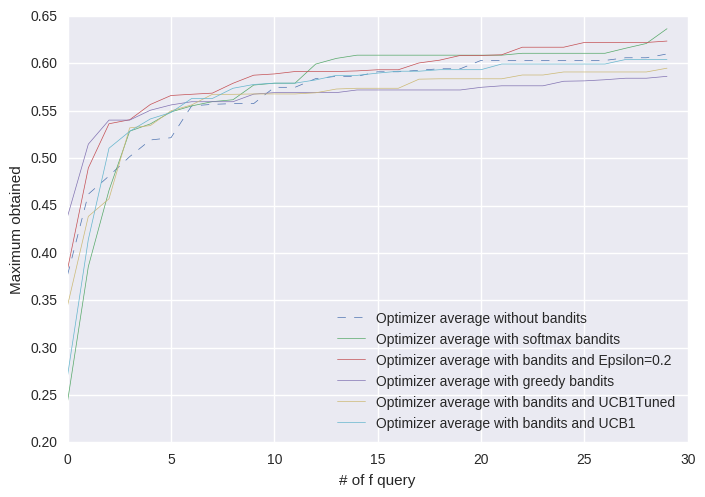}
  \caption{Comparison with the different average convergence curves for the success metric optimizers with different bandit algorithms.}
  \label{fig:clei_success_vs_bandits}
\end{figure}

The results of this experiment on the success metric variant optimizer are summarized in Table \ref{table:execution_times_of_different_optimizer_variants_success} for the total amount of queries on $f(\theta)$ and execution times, and in Fig \ref{fig:clei_success_vs_bandits} learning curves are shown. The average of the convergence of each variant with its corresponding bandit algorithm is presented in continuous curves, while the average convergence without bandits (as was mentioned in Section \ref{subsect:optimizers_convergence}) is displayed as a non-continuous curve. It can be appreciated, in terms of the number of queries made, how the bandit-based approach achieves a significant saving, between 15\% and 40\%, depending on the algorithm used. More specifically, it is worth noting that the $\epsilon$-greedy bandit and, specially, the Softmax bandit gives rise to the biggest reduction of all in the average amount of queries and elapsed time, followed by the greedy algorithm. On the other hand, the UCB1 and UCB1Tuned algorithms correspond to the lowest performance. By observing the learning curves, it can be seen how, in terms of convergence to the maximum, the variant with $\epsilon$-greedy bandit algorithm converges slightly better than the performance average with no bandits by an amount of 3,1\%, whereas Softmax starts with a poor convergence and then recovered to an average of 0.16\% increase. Finally, the greedy bandit exhibits a similar convergence (-0.1\%) and, regarding the UCB1 and UCB1Tuned algorithms, they slightly underperformed in comparison to the average performance without bandits (-0.48\% and -1.46\%, respectively).

\begin{table}
\centering
\caption{Execution times and average number of queries of different optimizer variants (success metric).}
\label{table:execution_times_of_different_optimizer_variants_success}
\begin{tabular}{@{}lcccccc@{}}
\toprule
                                             & No bandits & Softmax                       & $\epsilon$-greedy             & Greedy                        & UCB1                          & UCB1Tuned \\ \midrule
Avg. number of queries & \multicolumn{1}{c}{150.0}     & \multicolumn{1}{c}{83.6}    & \multicolumn{1}{c}{97.1}     & \multicolumn{1}{c}{115.9}     & \multicolumn{1}{c}{126.4}    & \multicolumn{1}{c}{125.4}    \\ \midrule
Avg. time per optimizer execution                   & \multicolumn{1}{c}{2:12:25}  & \multicolumn{1}{c}{1:19:47} & \multicolumn{1}{c}{1:27:33} & \multicolumn{1}{c}{1:44:48} & \multicolumn{1}{c}{1:54:37} & \multicolumn{1}{c}{1:52:53} \\ \midrule
\% of time reduction                               & -                              & 39.7\%                       & 33.8\%                       & 20.8\%                      & 13.4\%                      & 14.7\%                       \\ \bottomrule
\end{tabular}
\end{table}

\begin{figure}
 \centering
 \includegraphics[width=0.60\textwidth]{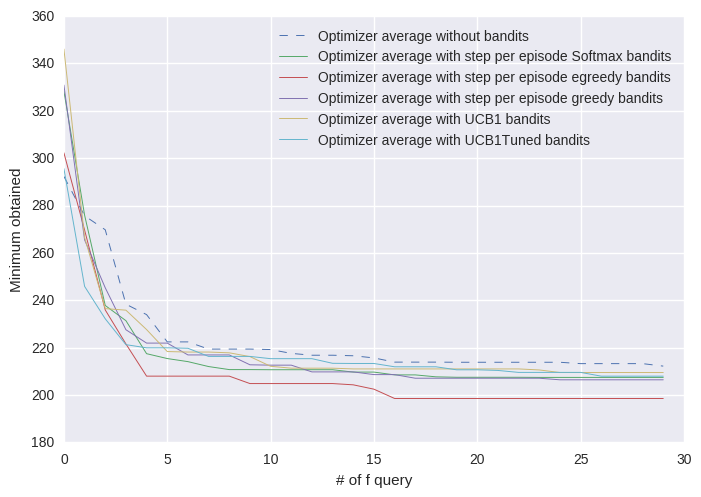}
  \caption{Comparison with the different average convergence curves for the step-per-episode measured optimizers with different bandit algorithms.}
  \label{fig:clei_stpep_vs_bandits}
\end{figure}

Finally, for the step-per-episode measured optimizers, the results are shown in Table \ref{table:execution_times_of_different_optimizer_variants_stpep} and in Fig. \ref{fig:clei_stpep_vs_bandits}. Regarding the number of queries made, it is shown that the $\epsilon$-greedy, Softmax and greedy algorithms were those that reduced the most the number of queries, and the UCB and UCB1Tuned obtained the least reduction, similarly to the success metric variant. The main differences in this experiment compared with the previous one is that the algorithm that achieves the biggest reduction was the $\epsilon$-greedy rather than Softmax. Moreover, this time the greedy algorithm performed consistently better. In terms of execution time reduction, $\epsilon$-greedy obtained the highest reduction with near 46\%, followed by a 40\% reduction of the greedy algorithm (which almost doubles the reduction obtained in the previous experiment). A slightly less significant reduction (39.4\%) is obtained using Softmax, which maintains almost the same reduction achieved in the previous experiment. In terms of convergence, all the bandit algorithms tried follow a better convergence, on average, towards the minimum, reporting less amount of steps-per-episode in comparison with the variant without bandits. Here, the $\epsilon$-greedy algorithm achieves best convergence, which found a minimum that is -6,4\% on average, compared with the optimizer variant that does not use bandits, and followed by Softmax, greedy, UCB1Tuned and UCB1 (-2.83\%, -2.63\%, -2.51\% and -1.48\%, respectively).

\begin{table}
\centering
\caption{Execution times and average number of queries of different optimizer variants (step-per-episode metric).}
\label{table:execution_times_of_different_optimizer_variants_stpep}
\begin{tabular}{@{}lcccccc@{}}
\toprule
                                             & No bandits & Softmax                       & $\epsilon$-greedy             & Greedy                        & UCB1                          & UCB1Tuned \\ \midrule
Avg. number of queries & \multicolumn{1}{c}{150.0}     & \multicolumn{1}{c}{89.4}    & \multicolumn{1}{c}{83.8}     & \multicolumn{1}{c}{93.3}     & \multicolumn{1}{c}{127.3}    & \multicolumn{1}{c}{127.9}    \\ \midrule
Avg. time per optimizer execution                   & \multicolumn{1}{c}{2:39:23}  & \multicolumn{1}{c}{1:36:28} & \multicolumn{1}{c}{1:26:36} & \multicolumn{1}{c}{1:34:41} & \multicolumn{1}{c}{2:15:29} & \multicolumn{1}{c}{2:14:09} \\ \midrule
\% of time reduction                               & -                              & 39.4\%                       & 45.6\%                       & 40.5\%                      & 14.9\%                      & 15.8\%                       \\ \bottomrule
\end{tabular}
\end{table}

Analyzing both experiments with and without bandits-based algorithms, it can be said that the main benefit of adding a bandit algorithm in the meta-learning layer is that, depending of the chosen algorithm, there is a considerable reduction regarding the number of queries to the performance function $f$. As a result,the execution time that every optimizer requires to complete their corresponding number of meta-episodes is also reduced. Regarding the convergence to the optimum, it is seen that some algorithms achieve, on average, a slight better performance than the variant with no bandits, while other algorithms obtained virtually the same convergence rate or even under-performed slightly. A problem to be addressed in a future work is the reliance of bandit algorithms on their own hyper-parameters; this is clearly seen in the greedy algorithm, which is the $\epsilon$-greedy algorithm with $\epsilon=0$, which performed consistently worse than the $\epsilon=0.2$ algorithm in both experiments. This is a plausibe problem that UCB1 and UCB1Tuned also had in their respective experiments. Even though they do not have hyper-parameters, the numeric scales can be considerably different than those of the problem as it happens in the experiments made. Thus, having a limited amount of queries, adding a hyper-parameter to change the scales can be an addition to be considered. Another aspect to address is that, even with the inclusion of bandit algorithms, the stagnation in the convergence it is repeated as in the variants without bandits, as it can be seen in the convergence curves. It can be said that the convergence to the maximum starts rapidly in the meta-episodes 2-5 (5.7\% increase and 4.77\% decrease per meta-episode for the different success and step-per-episode variants, in average), slowing down to an average of about 0.45\%  increase and 0.24\% decrease from there, respectively. As in the first set of experiments, this means that it would be almost unnoticeable in terms of convergence of the learning if the optimizer were run for half or even less meta-episodes.

\section{Concluding Remarks} \label{sect:conclusion}

A framework for automating the quest for a near-optimal policy in an autonomous reinforcement learning task has been presented. The proposed framework allows the learning agent to automatically find a good policy abstracting the user from having to manually tune the configuration, resorting to an expensive trial-and-error approach or using a possibly sub-optimal, default hyper-parameter configuration not suited for the specifics of the learning task at hand. By using Bayesian optimization, the framework takes into account all the previous points and observed values. By assuming a Gaussian prior of the data at the meta-level of learning, the framework uses a Gaussian process regression model to estimate the point where the maximum cumulative reward has a greater probability of being located, thus increasing the efficiency of the search of the hyper-parameter configuration that maximizes the efficiency of a reinforcement learning agent. Finally, by incorporating a bandit learning algorithm in order to decide which regions are potentially more appealling to explore further in order to reduce the stochasticity of the samples, the framework can reduce the amount of computational time required while maintaining a good convergence towards the optimum.

For future research work, there are several directions were this work is to be extended. The first direction points towards the extension of the performance functions and the respective analysis and comparison with the current ones in different environments and including metrics with labels that belongs to a certain category $C_i \in \{C_1, C_2, ..., C_n\}$ e.g. whether an execution is a failure or a success, instead of being real valued. The second direction is about the extension of the case study to an industrial example for a rescheduling task, in order to analyze how the framework performs with the increase of complexity in the environment, adapting the framework to such complexity by making modifications such as the use of more complex reinforcement learning algorithms or erasing queries that have low information contents after the covariance matrix reaches a certain rank. Also, another research avenue is the extension of the framework to a higher level for hyper-parameter selection such as RL algorithm selection and agent policy selection, among others. Additionally, other research direction is to add in the framework methods to detect which regions of hyper-parameters can ensure an adequate performance in the agent's learned policy, and then perform an optimization inside those smaller regions. The fifth direction address the incorporation in RLOpt of the optimization of the Gaussian process own hyper-parameters, by maximizing Eq. \eqref{eqn:log_likelihood} instead of using empirically determined values. Finally, the sixth direction approaches the implementation of other regression models such as random forests, and the comparison of GP with different covariance functions.

\section{Acknowledgement}

This work was supported by the National Technological University of Argentina (UTN) and the National Scientific and Technical Research Council of Argentina (CONICET), and by the projects UTI4375TC and EIUTNVM0003581, both funded by the UTN.

\bibliography{Zotero.bib}

\end{document}